# Convergent Message-Passing Algorithms for Inference over General Graphs with Convex Free Energies

Tamir Hazan     Amnon Shashua

School of Eng. & Computer Science
The Hebrew University of Jerusalem
Israel

## Abstract

Inference problems in graphical models can be represented as a constrained optimization of a free energy function. It is known that when the Bethe free energy is used, the fixed-points of the belief propagation (BP) algorithm correspond to the local minima of the free energy. However BP fails to converge in many cases of interest. Moreover, the Bethe free energy is non-convex for graphical models with cycles thus introducing great difficulty in deriving efficient algorithms for finding local minima of the free energy for general graphs. In this paper we introduce two efficient BP-like algorithms, one sequential and the other parallel, that are guaranteed to converge to the global minimum, for any graph, over the class of energies known as "convex free energies". In addition, we propose an efficient heuristic for setting the parameters of the convex free energy based on the structure of the graph.

## 1 Introduction

Probabilistic graphical models present a convenient and popular tool for reasoning about complex distributions. The graphical model reflects the way the complex distribution factors into a product of distributions over a small number of variables (cliques), where the graph represents the incidence between cliques and the variables contained in them — such a graph is known as a *factor graph*. The probabilistic inference is represented by a way of calculating marginal distributions (or the most likely assignment of variables) efficiently using the structure of the graph.

One of the most popular class of methods for inference over (factor) graphs are message-passing algorithms which pass messages along the edges of the factor graph until convergence. The belief propagation (BP) algorithm (and its extensions) is popular, and has received the most attention, due to its simplicity and computational efficiency. The BP algorithm is exact, i.e., the resulting marginal distributions are the correct ones, when the factor graph is free of cycles. An intriguing feature of BP, which most likely is the source for its great popularity, is that it often gives surprisingly good approximate results for graphical models with cycles. However, in this context there are no convergence guarantees (except under some special cases (Mooij & Kappen, 2005)) and the algorithm often fails to converge.

It is known that the fixed-points of the BP algorithm correspond to local minima of a constrained energy function called the Bethe free energy (Yedidia et al., 2005). The free energy arises from the expansion of the KL-divergence between the input distribution and its product form. The Bethe function replaces the entropy term in the free energy by an approximation which is exact for factor graphs without cycles. In such a case, the Bethe free energy is convex over the set of constraints (representing validity of marginals). When the factor graph has cycles the Bethe energy is non-convex and although it is possible to derive convergent algorithms to a local minima of the Bethe function (Yuille, 2002) the computational cost is large and thus has not gained popularity.

To overcome the difficulty with the non-convexity of the Bethe approximation, several authors have introduced a class of approximations known as *convex free energies* which are convex over the set of constraints for any factor graph. An important member of this class is the tree-reweighted (TRW) free energy which consist of a linear combination of free energies defined on spanning trees of the factor graph. It is notable that for this specific member of convex free energies a convergent message-passing algorithm has been recently introduced (Globerson & Jaakkola, 2007b). The algorithm is sequential (unlike BP which has both se-quential and parallel forms) and applies to graphs with pairwise cliques only. However, a convergent message passing algorithm for the general class of convex free energies is still lacking. The existing algorithms either employ damping heuristics to ensure convergence in practice (Wainwright et al., 2005) or focus on a sub-class of free energies where the entropy term is a positive combination of joint entropies (Heskes, 2006).

In this paper, we derive convergent message-passing algorithms, one sequential and the other parallel, for the general class of convex free energies. The derivation applies to general factor graphs (cliques of any size) and have a similar architecture to the BP algorithm. The algorithms are based on a general framework for handling optimization problems of the type $f(b) + \sum_i h_i(b)$ where $f(b)$ is strictly convex and $h_i(b)$ are convex but *not necessarily strict nor differentiable*. We show that problems of this class have a simple block-update (sequential and parallel) message-passing solution. We then map the constrained convex free energy problem into this framework.

Independently, we propose also a heuristic for setting up the parameters of the convex free energy from the structure of the factor graph. The key idea is to strive for a convex free energy which is as close as possible to Bethe's free energy under a set of constraints governing the class of convex free energies. The underlying motivation is borne by the empirical observation from BP practitioners that when BP does converge, the results are often surprisingly good (Murphy et al., 1999). Since our scheme would always converge and the free energy approximation is close to Bethe's, we would have in some sense a "convergent BP" for general graphs.

## 2 Terminology and Problem Setup

We consider a joint distribution $P(\mathbf{x})$ on a set of discrete variables $\mathbf{x} = x_1, ..., x_n$ in a finite domain. In a graphical model we suppose that $P(\mathbf{x})$ factors into a product of non-negative functions (clique potentials):

$$P(\mathbf{x}) = \frac{1}{Z} \prod_\alpha \psi_\alpha(\mathbf{x}_\alpha),$$

where $\alpha$ is an index labeling $m$ functions $\psi_1, ..., \psi_m$ and the function $\psi_\alpha(\mathbf{x}_\alpha)$ has arguments $\mathbf{x}_\alpha$ that are some subset of $\{x_1, ..., x_n\}$ and $Z$ is a normalization constant. The factorization structure above is conveniently represented by a *factor graph* (Kschischang et al., 2001) which is a bipartite graph with variable nodes one for each variable $x_i$ and a factor node for each function $\psi_\alpha$. An edge connects a variable node $i$ with factor node $\alpha$ if and only if $x_i \in \mathbf{x}_\alpha$, i.e., $x_i$ is an argument of $\psi_\alpha$. We adopt the terminology where

$N(i)$ stands for all factor nodes that are neighbors of variable node $i$ and $N(\alpha)$ stands for all variable nodes that are neighbors of factor node $\alpha$. Finally, we limit our treatment to factor graphs where any pair of factors intersect in at most a single variable node, i.e., $N(\alpha) \cap N(\beta)$ is either empty or equal to some $i$. There is no technical limitation to allow for higher-order intersections but that comes at the price of reducing the clarity of our presentation. In Section 7 we will explain the necessary additions for handling general factor graphs.

The typical task we try to perform is to compute the marginal distributions $P(x_i) = \sum_{\mathbf{X} \setminus x_i} P(\mathbf{x})$ and $P(\mathbf{x}_\alpha) = \sum_{\mathbf{X} \setminus x_\alpha} P(\mathbf{x})$. Basically, the computation requires the summation over the states of all the variable nodes not in $\mathbf{x}_\alpha$ (including the case of the singleton $x_i$). This computation is generally hard because it can require summing up exponentially large number of terms — thus one seeks efficient ways or approximate solutions for the marginals.

Let $b(\mathbf{x})$ stand for the approximate distribution where $b_i(x_i)$ approximates $P(x_i)$ and $b_\alpha(\mathbf{x}_\alpha)$ approximates $P(\mathbf{x}_\alpha)$ with the constraints that $\sum_{\mathbf{x}_\alpha \setminus x_i} b_\alpha(\mathbf{x}_\alpha) = b_i(x_i)$ for all $\alpha \in N(i)$. The *free energy* arises from minimizing the KL-divergence between the approximate distribution $b(\mathbf{x})$ and the un-normalized product form:

$$D(b \,\|\, \prod_\alpha \psi_\alpha(\mathbf{x}_\alpha)) = \sum_\alpha \sum_{\mathbf{x}_\alpha} E_\alpha(\mathbf{x}_\alpha) b_\alpha(\mathbf{x}_\alpha) - H(b),$$

where $E_\alpha = -\ln \psi_\alpha$ and $H(b)$ is the entropy of $b(\mathbf{x})$. In other words, the free energy consists of a sum of a linear term over $b$ which is exponential in the size of the cliques and and an entropy term (which is exponential in $n$). The approximate methods for computing the marginals are based on choosing an approximation to the entropy term $H(b)$.

The Bethe free energy approximates $H(b)$ by $\sum_\alpha H(b_\alpha) + \sum_i (1 - d_i) H(b_i)$ where $d_i$ is the degree of the variable node $i$, $H(b_\alpha) = -\sum_{\mathbf{x}_\alpha} b_\alpha(\mathbf{x}_\alpha) \ln b_\alpha(\mathbf{x}_\alpha)$ and $H(b_i) = -\sum_{x_i} b(x_i) \ln b(x_i)$. As a result, the computational complexity of the Bethe free energy is exponential only in the size of the cliques. The Bethe free energy is exact (equal to free energy) when the factor graph has no cycles and in that case the energy is strictly convex over the set of constraints mentioned above. When the factor graph has cycles the Bethe free energy is non-convex. Another notable property is that the fixed-points of the BP algorithm correspond to local minima of the Bethe free energy minimization over the constraints on $b$ (Yedidia et al., 2005).

The Bethe approximation of the entropy $H(b)$ can be

written in a more general form as:

$$\sum_\alpha \bar{c}_\alpha H(b_\alpha) + \sum_i \bar{c}_i H(b_i), \quad (1)$$

where $\bar{c}_i = 1 - \sum_{\alpha \in N(i)} \bar{c}_\alpha$. Thus when the coefficients $\bar{c}_\alpha = 1$ for all factor nodes we obtain the Bethe approximation. A *convex free energy* is based on a result of (Heskes, 2004) who derived sufficient conditions for an entropy approximation to be convex over the set of constraints. In the setting we have described, those conditions have the following form (Weiss et al., 2007):

**Definition:** An approximate entropy term of the form eqn. 1 is strictly convex over the set of constraints $\sum_{\mathbf{x}_\alpha \setminus x_i} b_\alpha(\mathbf{x}_\alpha) = b_i(x_i)$ for all $\alpha \in N(i)$ if there exists $c_i, c_{i\alpha} \geq 0$ and $c_\alpha > 0$ such that $\bar{c}_\alpha = c_\alpha + \sum_{i \in N(\alpha)} c_{i\alpha}$ and $\bar{c}_i = c_i - \sum_{\alpha \in N(i)} c_{i\alpha}$. The approximate entropy becomes:

$$\sum_{i,\alpha \in N(i)} c_{i\alpha}(H(b_\alpha) - H(b_i)) + \sum_\alpha c_\alpha H(b_\alpha) + \sum_i c_i H(b_i)$$

Taken together, the convex free energy constrained optimization problem is:

$$\min_{b_\alpha, b_i} \sum_\alpha \sum_{\mathbf{x}_\alpha} E_\alpha(\mathbf{x}_\alpha) b_\alpha(\mathbf{x}_\alpha) - \sum_\alpha c_\alpha H(b_\alpha) \quad (2)$$
$$- \sum_i c_i H(b_i) + \sum_{i,\alpha \in N(i)} c_{i\alpha}(H(b_i) - H(b_\alpha))$$

subject to
$$\sum_{\mathbf{x}_\alpha \setminus x_i} b_\alpha(\mathbf{x}_\alpha) = b_i(x_i) \quad \forall \alpha \in N(i)$$
$$\sum_{\mathbf{x}_\alpha} b_\alpha(\mathbf{x}_\alpha) = 1 \quad \forall \alpha$$
$$b_\alpha, b_i \geq 0$$

We denote the criterion function (convex free energy) by $F_{con}(b_\alpha, b_i)$ and note that it is strictly convex over the set of constraints provided that $c_i, c_{i\alpha} \geq 0$ and $c_\alpha > 0$. For now we assume that the parameters $c_i, c_{i\alpha}, c_\alpha$ are given as input and set out to derive a message-passing algorithm (two versions, one sequential and the other parallel) which is guaranteed to converge to the global minimum for any factor graph. Later in Section 5 we will introduce an algorithm for determining the convex free energy parameters from the factor graph.

## 3 A General Framework for Sequential and Parallel Message Passing Algorithms

The constrained minimization of eqn. 2 can be handled within the body of convex programming tools. Those,

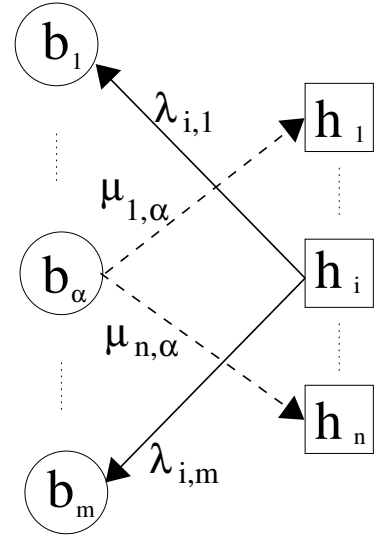

Figure 1: *Message-passing architecture of Algorithms 1,2*

however, have a high computational cost and their architecture (the update flow of parameters) is far from similar to a message passing architecture and to BP in particular. In this section we will take a detour and develop a message passing framework (sequential and parallel versions) to a particular sub-class of convex problems. Later in Section 4 we will map the convex free energy minimization of eqn. 2 into this framework. Consider the class of problems

$$\min_b f(b) + \sum_{i=1}^n h_i(b),$$

where $b \in R^m$, $f(b)$ is a strictly convex real valued function and $h_i(b)$ are convex (not necessarily strict nor differentiable), and *proper* (i.e., can take the value $\infty$ for some values of $b$). This class of problems includes in particular the classical convex program: $\min_b f(b)$ over the constraints $b \in C_1 \cap \cdots \cap C_n$ where $C_i$ are convex sets[1].

For this class of problems we derive two message-passing algorithms, one sequential and the other parallel. Both algorithms are based on a block update regime using convex duality. The sequential algorithm is described below:

**Algorithm 1 (Sequential Message-Passing)** *Let $\boldsymbol{\lambda}_i$ and $\boldsymbol{\mu}_i$, $i = 1, ..., n$ be vectors in $R^m$. Set $\boldsymbol{\lambda}_i = 0$.*

1. *For $t = 1, 2, ...$*

2. *For $i = 1, ...n$:*

    (a) $\boldsymbol{\mu}_i \leftarrow \sum_{j \neq i} \boldsymbol{\lambda}_j$

---
[1] In this case $h_i(b) = \delta_{C_i}(b)$ is the indicator function $\delta_{C_i}(b) = 0$ if $b \in C_i$ and $\delta_{C_i}(b) = \infty$ if $b \notin C_i$.

(b) $\mathbf{b}^* \leftarrow \underset{\mathbf{b} \in domain(h_i)}{\operatorname{argmin}} \left\{ h_i(\mathbf{b}) + f(\mathbf{b}) + \mathbf{b}^\top \boldsymbol{\mu}_i \right\}.$

(c) $\boldsymbol{\lambda}_i \leftarrow -\boldsymbol{\mu}_i - \nabla f(\mathbf{b}^*).$

Output $\mathbf{b}^*$.

The vectors $\boldsymbol{\lambda}_i$ and $\boldsymbol{\mu}_i$ are messages passed along edges of a bipartite graph with $n$ (function) nodes corresponding to the $n$ functions $h_i(b)$ and $m$ (variable) nodes corresponding to the dimension of $b$. Function node $i$ sends the $m$ coordinates of vector $\boldsymbol{\lambda}_i$ to the $m$ variable nodes. Variable node $j$ sends the $j$'th coordinate of vectors $\boldsymbol{\mu}_1, ..., \boldsymbol{\mu}_n$ to the $n$ functions nodes (see Fig. 1). The algorithm is "sequential" in the sense that it is crucial to move sequentially over the index $i = 1, ..., n$, thus the network proceeds in a node-after-node update policy.

For those familiar with successive projection schemes, in the particular case when $h_i(b) = \delta_{C_i}(b)$ (the indicator function of convex set $C_i$), the update step for $b$ is a "Bregman" projection (Bregman et al., 1999) of the vector $\boldsymbol{\mu}_i$ onto the convex set $C_i$. In that case, following some algebraic manipulations (such as eliminating $\boldsymbol{\mu}_i$ among other manipulations) the scheme reduces to the well known Dykstra (Dykstra, 1983) (also goes under different names such as Hildreth, Bregman, Csiszar, Han, Tseng) successive projection algorithm which has its origins in the work of Von-Neumann (von Neumann, 1950). We introduce next a *parallel* message-passing algorithm:

**Algorithm 2 (Parallel Message-Passing)** Let $\boldsymbol{\lambda}_i$ and $\boldsymbol{\mu}_i$, $i = 1, ..., n$ be vectors in $R^m$. Set $\boldsymbol{\mu}_i = 0$.

1. For $t = 1, 2, ...$

2. For $i = 1, ...n$ in parallel

$$\mathbf{b}^* \leftarrow \underset{\mathbf{b} \in domain(h_i)}{\operatorname{argmin}} \left\{ \frac{1}{n} f(\mathbf{b}) + h_i(\mathbf{b}) + \mathbf{b}^\top \boldsymbol{\mu}_i \right\}$$

$$\boldsymbol{\lambda}_i \leftarrow -\boldsymbol{\mu}_i - \frac{1}{n} \nabla f(\mathbf{b}^*)$$

3. For $i = 1, ...n$ in parallel

$$\boldsymbol{\mu}_i \leftarrow -\boldsymbol{\lambda}_i + \frac{1}{n} \sum_{j=1}^k \boldsymbol{\lambda}_j$$

Output $\mathbf{b}^*$.

The description of the messages and the network architecture are the same as in the sequential algorithm but here all the function nodes update and send their messages in parallel to the variable nodes. Once all the variable nodes have received their messages they compute their update and send their message in parallel to the function nodes. The derivation of both algorithms is presented in the Appendix.

## 4 Convergent Message-Passing Algorithms for Convex Free Energies

We are ready to derive a convergent algorithm for the constrained convex free energy minimization problem (eqn. 2) using the two algorithms above. It is important to note that in the framework of $f(b) + \sum_i h_i(b)$ the function $f(b)$ is strictly convex in the entire domain whereas the convex free energy $F_{con}(b_\alpha, b_i)$ is strictly convex *over the set of constraints*. We need therefore both to map $F_{con}(b_\alpha, b_i)$ onto the framework of $f(b) + \sum_i h_i(b)$ and handle the convexity over the domain of definition issue.

Let $b_\alpha(x_i)$ stand for $\sum_{\mathbf{x}_\alpha \setminus x_i} b_\alpha(\mathbf{x}_\alpha)$ (a notation also used by (Heskes, 2006)). The marginal constraints dictate that $b_\alpha(x_i) = b_\beta(x_i)$ for all $\alpha, \beta \in N(i)$. We substitute $b_\alpha(x_i)$ for $b_i(x_i)$ in eqn. 2 and move terms around to fit the $f(b) + \sum_i h_i(b)$ framework. The result is summarized below:

Let $b = (b_{\alpha_1}, ..., b_{\alpha_m})$, that is we dropped $b_i(x_i)$ from the process. Let $f(b)$ be defined as follows:

$$f(b) = \sum_\alpha \left( \sum_{x_\alpha} E_\alpha(x_\alpha) b_\alpha(x_\alpha) - c_\alpha H_\alpha(b_\alpha) \right) \quad (3)$$
$$= \sum_\alpha f_\alpha(b_\alpha)$$

Let $h_i(b)$, $i = 1, ..., n$, corresponding to the $n$ variable nodes of the factor graph, be defined below:

$$h_i(\mathbf{b}) = \left\{ \begin{array}{ll} \infty & \exists \alpha, \beta \in N(i) : b_\alpha(x_i) \neq b_\beta(x_i) \\ \sum_{\alpha \in N(i)} h_{i\alpha}(b_\alpha) & otherwise \end{array} \right\} \quad (4)$$

where $h_{i\alpha}(b_\alpha)$ is defined as follows:

$$h_{i\alpha}(b_\alpha) = (c_i/|N(i)| - c_{i\alpha}) \sum_{x_i} b_\alpha(x_i) \ln b_\alpha(x_i) - c_{i\alpha} H(b_\alpha) \quad (5)$$

Taken together, the constrained convex free-energy minimization (eqn. 2) becomes:

$$\min_{b=(b_{\alpha_1},...,b_{\alpha_m})} f(b) + \sum_{i=1}^n h_i(b) \quad s.t. \quad b \geq 0, \sum_{x_\alpha} b_\alpha(x_\alpha) = 1,$$

where now $f(b)$ is strictly convex over its entire domain of definition and $h_i$ are proper convex functions. Note that $f(b)$ and $h_i(b)$ each decompose into a sum of simpler functions, i.e., $f(b) = \sum_\alpha f_\alpha(b_\alpha)$ and $h_i(b) = \sum_{\alpha \in N(i)} h_{i\alpha}(b_\alpha)$. The decomposition translates into the messages being sparse. For example, the update step of $b^*$ in both message-passing algorithms becomes:

$$b^*_{i,\alpha \in N(i)} = \underset{\mathbf{b} \in dom(h_i)}{\operatorname{argmin}} \sum_{\alpha \in N(i)} (f_\alpha(b_\alpha) + h_{i\alpha}(b_\alpha) + b_\alpha^\top \boldsymbol{\mu}_{i\alpha})$$

1. Set $n_{j\to\gamma}(x_\gamma) = 1$ for all $i = 1, ..., n$, $\gamma \in N(i)$ and $\mathbf{x}_\gamma$.

2. For $t = 1, 2, ...$

3. For $i = 1, ...n$:

$$m_{\gamma\to i}(x_i) = \sum_{z_\gamma \setminus x_i} \left( \psi_\gamma(z_\gamma) \prod_{j\in N(\gamma)\setminus i} n_{j\to\gamma}(z_\gamma) \right)^{1/\hat{c}_{i\gamma}}$$

$$b^*_\gamma(x_i) \propto \prod_{\alpha\in N(i)} m^{\hat{c}_{i\alpha}/\hat{c}_i}_{\alpha\to i}(x_i),$$

$$n_{i\to\gamma}(x_\gamma) = \left( \psi_\gamma(x_\gamma) \prod_{j\in N(\gamma)\setminus i} n_{j\to\gamma}(x_\gamma) \right)^{-\frac{c_{i\gamma}}{\hat{c}_{i\gamma}}} \left( \frac{b^*_\gamma(x_i)}{m_{\gamma\to i}(x_i)} \right)^{c_\gamma}$$

Figure 2: Sequential message-passing algorithm for convex free energy. The constants $\hat{c}_i$ and $\hat{c}_{i\alpha}$ are defined as $\hat{c}_i = c_i + \sum_{\alpha\in N(i)} c_\alpha$ and $\hat{c}_{i\alpha} = c_\alpha + c_{i\alpha}$.

where $b_{i,\alpha\in N(i)}$ are the entries $b_\alpha$ in $b$ corresponding to factor nodes $\alpha$ neighboring to variable node $i$. Likewise $\boldsymbol{\mu}_{i\alpha}$ are the portions in $\boldsymbol{\mu}_i$ corresponding to factor nodes neighboring to variable node $i$. The domain of $h_i$ are the constraints :

$$\sum_{\mathbf{x}_\alpha} b_\alpha(\mathbf{x}_\alpha) = 1, \quad \forall \alpha \in N(i)$$
$$b_\alpha(x_i) = b_\beta(x_i), \quad \forall \alpha, \beta \in N(i)$$

We omit the remainder of the derivation as it is long, tedious and mechanical and arrive to the final description of the two message-passing algorithms presented in Fig. 2 and Fig. 3. Like BP, the algorithms send messages between variable nodes and factor nodes where $n_{i\to\gamma}(x_\gamma)$ represents the message from variable node $i$ to factor node $\gamma$ and $m_{\gamma\to i}(x_\gamma)$ is the message from factor node $\gamma$ to variable node $i$.

## 5 Fitting a Convex Free Energy to a Graphical Model

The convex free energy contains three sets of parameters $c_i, c_{i\alpha} \geq 0$ and $c_\alpha > 0$ one parameter per each node and edge in the factor graph. Our discussion so far was general in the sense that we presented algorithms for handling the family of convex free energies without regard as to how those parameters are determined. The only setting of parameters proposed to date is the tree-reweighted (TRW) free energy where $\bar{c}_\alpha$ can be set analytically. This has a simple form when the cliques are of size 2, i.e., representing pairs of variables. In that case, $\bar{c}_\alpha$ is the number of spanning trees containing $\alpha$ divided by the total number of spanning trees (a computation which can be done analytically). Once $\bar{c}_\alpha$ is determined then $\bar{c}_i = 1 - \sum_{\alpha\in N(i)} \bar{c}_\alpha$ and from $\bar{c}_\alpha$ and $\bar{c}_i$ one can solve for $c_i, c_{i\alpha} \geq 0$ and $c_\alpha > 0$ by means of linear satisfaction from the equations:

$$\bar{c}_\alpha = c_\alpha + \sum_{i\in N(\alpha} c_{i\alpha}, \quad \bar{c}_i = c_i - \sum_{\alpha\in N(i)} c_{i\alpha}. \quad (6)$$

In this section we propose a heuristic for setting the parameters based on the following idea[2]. Given the equations (eqn. 6) connecting the parameters to $\bar{c}_\alpha$ and $\bar{c}_i$, the space of admissible solutions must satisfy the following equations:

$$c_i + \sum_{\alpha\in N(i)} (c_\alpha + \sum_{j\in N(\alpha)\setminus i} c_{j\alpha}) = 1, \quad i = 1, ..., n$$
$$c_i, c_{i\alpha} \geq 0, \quad c_\alpha > 0.$$

Among all possible admissible solutions we choose the one in which $\bar{c}_\alpha$ is as uniform as possible, i.e., we apply Laplace's principle of insufficient reasoning. The criterion function, therefore, minimizes:

$$\min_{c_i, c_{i\alpha}, c_\alpha \in admissible} \sum_\alpha (c_\alpha + \sum_{i\in N(\alpha)} c_{i\alpha} - 1)^2, \quad (7)$$

which is a least-squares criteria for uniformity of $\bar{c}_\alpha$. Alternatively, we also used the maximum entropy approach where the criterion function minimizes $\sum_\alpha \bar{c}_\alpha \ln \bar{c}_\alpha$. In both cases we used standard solvers to recover $c_i, c_{i\alpha}, c_\alpha$, i.e., we did not attempt to devise specifically tailored solvers for those problems.

The desire towards uniformity, besides being used extensively in probabilistic settings, is motivated by the success of the Bethe free energy where $\bar{c}_\alpha = 1$. The Bethe free energy is non-convex for factor graphs with cycles, thus is not a member of the convex free energies, but empirical evidence suggest that when BP converges the marginals are surprisingly good. For Bethe free energy $\bar{c}_\alpha = 1$ over all factor nodes $\alpha$ — hence our proposal to strive for uniformity over the space of admissible solutions. In some sense we are attempting to "convexify" the Bethe free energy, although this is not being done directly.

---
[2] A similar idea was independently derived by Nir Friedman and his collaborators — personal communication.

## 6 Experiments

We applied our (parallel and sequential) message passing algorithm using the heuristic (eqn. 7) for setting the parameters of the convex free energy from the input graph to an Ising model on a two dimensional $8 \times 8$ grid. The distribution has the form $p(\mathbf{x}) \propto e^{\sum_{ij \in E} \theta_{ij} x_i x_j + \theta_i x_i}$, where $\theta_{ij}, \theta_i$ are parameters, $x_i \in \{\pm 1\}$, and $E$ are edges of the 2D grid.

Following (Globerson & Jaakkola, 2007a), the parameters $\theta_i$ were drawn uniformly from $\mathcal{U}[-d_f, d_f]$ where $d_f \in \{0.05, 1\}$. The parameters $\theta_{ij}$ were drawn from $\mathcal{U}[-d_o, d_o]$ or $\mathcal{U}[0, d_o]$ to obtain *mixed* or *attractive* interaction potential respectively. The interaction levels were $d_o \in \{0.2, 0.4, ..., 4\}$. In addition to BP[3], the following algorithms were used to estimate the marginals of the distribution:

- CCCP algorithm (Yuille, 2002) for obtaining a local minima of the constrained Bethe free energy. Our message-passing algorithm runs on a "covexified" version of the Bethe free-energy and achieves its global minimum. The CCCP, on the other hand, runs on the (non-convex) Bethe energy and finds a local minima.

- Sequential MP where the convex free energy parameters determined using the convex-$L_2$ free energy described in Eqn.7.

- The same as above but the parameters were set using maximum entropy (instead of $L_2$), we call convex-$H$.

- The TRW method of (Wainwright et al., 2005) with uniform distributions over the trees. Since the TRW free energy belongs to the class of convex free energies we ran our sequential MP algorithm on the parameters $c_i, c_{i\alpha}, c_\alpha$ determined by TRW.

For each setting of the parameters and each algorithm we calculated the mean $L_1$ error in the marginals $\frac{1}{n}\sum_i |p^{(alg)}(x_i = 1) - p^{(true)}(x_i = 1)|$. The accuracy

---

[3] We used the inference package by Talya Meltzer available at http://www.cs.huji.ac.il/~talyam/.

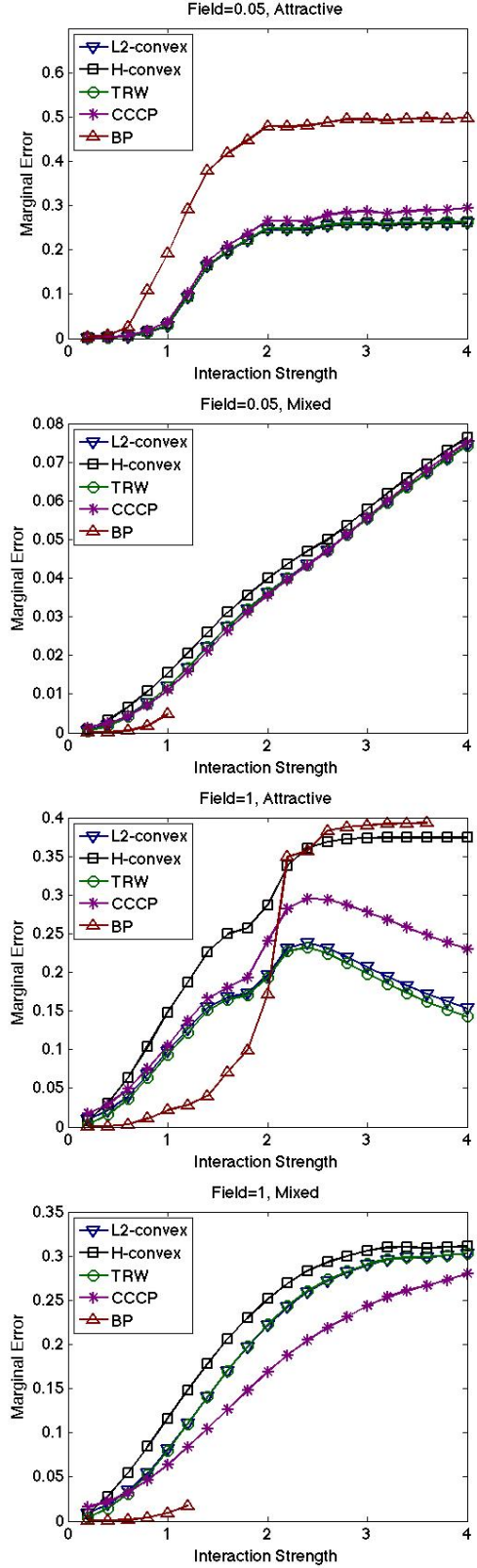

Figure 4: *Comparison of error in marginals estimation on an Ising model on a two dimensional $8 \times 8$ grid. The models presented include BP (when converged), TRW, CCCP, convex-$L_2$ and convex-$H$. Mean is shown for 10 random trials.*

1. For $t = 1, 2, ...$

2. For $i = 1, ...n$ in parallel

$$m_{\gamma \to i}(x_i) = \sum_{z_\gamma \setminus x_i} \left( \psi_\gamma(z_\gamma) \frac{\prod_{j \in N(\gamma)} n_{j \to \gamma}(z_\gamma)^{1/n}}{n_{i \to \gamma}(z_\gamma)} \right)^{1/\hat{c}_{i\gamma}}$$

3. For $i = 1, ...n$ in parallel

   (a) For every $x_i$:

   $$b^*_\gamma(x_i) \propto \prod_{\alpha \in N(i)} m_{\alpha \to i}^{\hat{c}_{i\alpha}/\hat{c}_i}(x_i)$$

   for every $\gamma \in N(i)$ and every $x_\gamma \setminus x_i$:

   $$n_{i \to \gamma}(x_\gamma) = \frac{n_{i \to \gamma}(x_\gamma)}{(\psi_\gamma(x_\gamma) \prod_{j \in N(\gamma)} n_{j \to \gamma}(x_\gamma))^{1/n}} \left( \psi_\gamma(z_\gamma) \frac{\prod_{j \in N(\gamma)} n_{j \to \gamma}(x_\gamma)^{1/n}}{n_{i \to \gamma}(x_\gamma)} \right)^{c_\gamma/n\hat{c}_{i\gamma}} \left( \frac{b^*_\gamma(x_i)}{m_{\gamma \to i}(x_i)} \right)^{c_\gamma/n}$$

Figure 3: Parallel message-passing algorithm for convex free energy. The constants $\hat{c}_i$ and $\hat{c}_{i\alpha}$ are defined as $\hat{c}_i = c_i + \sum_{\alpha \in N(i)} c_\alpha$ and $\hat{c}_{i\alpha} = c_\alpha + c_{i\alpha}$.

results are shown in Fig. 4. The displays are arranged into four cases: Field=0.05, 1 and Mixed versus Attractive interactions. In three out of the four cases, the performance of the three convex free energies models are roughly the same. In the case [Field=1, Attractive], TRW and convex-$L_2$ produce roughly the same marginal approximations and convex-$H$ is worse. BP does not converge in the Mixed cases for high Interaction value of $d_o$; CCCP produces comparable results in the two cases when Field=0.05, and produces the best results for [Field=1, Mixed]. To conclude so far, the two convex free energy settings produce comparable results to TRW on all cases and are sometimes better and sometimes worse than BP (when converges). Among the two setting of the convex free energy, convex-$L_2$ consistently produces better approximations than convex-$H$.

It is interesting that CCCP often produces good approximations, however this comes at a costly run-time tradeoff. Fig. 5 compares the running time of our (sequential) MP algorithm with a general convex solver performing conditional gradient descent on the primal energy function (Bertsekas et al., 2003) which uses linear programming to find feasible search directions, and to the CCCP algorithm. We ran all three algorithms on $n \times n$ grids where $n = 2, 3, ..., 10$. The stopping criteria for all algorithms was the same and based on a primal energy difference of $10^{-5}$. For a $10 \times 10$ grid, for instance, the general convex solver was slower by a factor of 20 (e.g., 306 seconds compared to 15.2) and the CCCP was slower by a factor of 115 compared to our MP algorithm (running 1767 seconds). For a $2 \times 2$ grid, on the other hand, our MP algorithm took 0.15

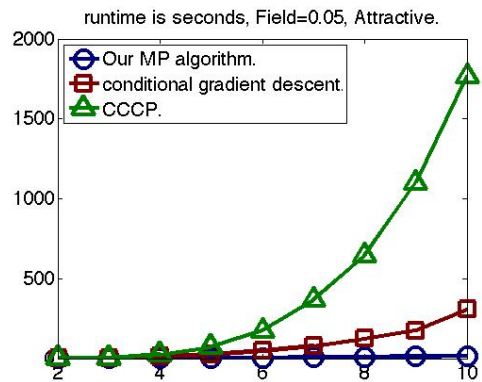

Figure 5: *Run-time (in seconds) comparisons of our message-passing algorithm against a conditional gradient descent solver (running on convex-$L_2$ free energy) and against CCCP for the (non-convex) Bethe energy. All three algorithms were applied to $n \times n$ grids with $n = 2, 3..., 10$. Mean is shown for 10 random trials.*

seconds compared to 0.59 for CCCP and 1.41 seconds for the general convex solver.

Our next experiment is conducted on random graphs to analyze the differences between BP, CCCP, TRW, convex-$H$ and convex-$L_2$. To generate a random graph we used the probability space of $G(n, p)$ over graphs with n vertices - where each edge is present with probability $p$ and absent with probability $1 - p$, independently among edges. Note that in $G(n, \frac{1}{2})$ all the graphs have the same probability, i.e., $G(n, \frac{1}{2})$ is the probability space consisting of all $n$-vertex graphs under the uniform distribution.

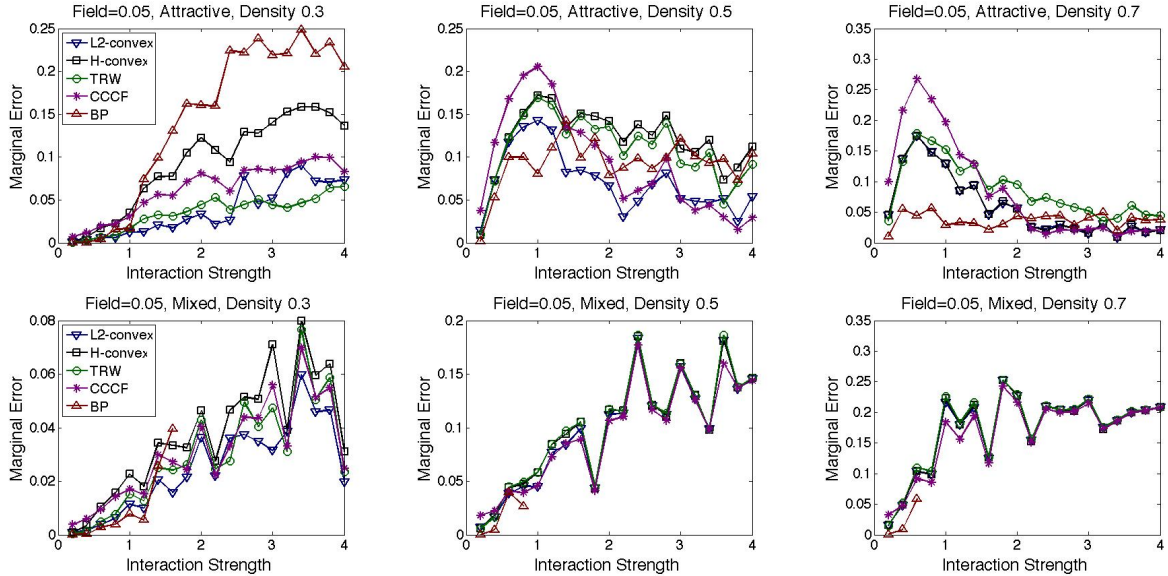

Figure 6: *Comparison of error in marginals estimation for random graphs with edge density $p = 0.3, 0.5, 0.7$ (left to right) and for local field value of $0.05$. For high field value ($d_f = 1$, not displayed here) convex-$L_2$, convex-H and TRW produce similar results. Mean is shown for $10$ random trials.*

In Fig. 6 we observe that the typical graph behavior is mainly dependent on the edge-density of the random graph as well as on the interaction levels. In order to compute the exact marginals we chose 10 vertices. A random graph with edge density of $p = 0.3$ is "almost" a tree and we see that the TRW is slightly inferior to convex-$L_2$ and better than convex-H. For intermediate edge-density $p = 0.5$ the TRW and convex-H are comparable and both are inferior to convex-$L_2$. When edge-density is high, i.e., $p = 0.7$ the graph is far from a tree and the convex-$L_2$ as well as convex-H are better than TRW. Note that in the Mixed case all the convex algorithms produce comparable results. In those cases BP usually does not converge. Nevertheless, in other cases (except $p = 0.3$, Attractive) BP produces good marginal approximation (consistent with empirical observations found in the literature) and that convex-$L_2$ is not far behind. It is interesting to note that, unlike the $8 \times 8$ grid, the CCCP is in most situations inferior to the convex free energy models.

## 7 Summary and Discussion

The convex free energies provide a way for obtaining approximate inference over general graphs. There are two main issues in this regard: the first is how to obtain a guaranteed globally convergent message-passing algorithm for the general class of convex free energies, and secondly, how to tune the energy parameters $c_i, c_{i\alpha}, c_\alpha$ to a specific graph?

As for the first issue, we have provided a complete treatment by deriving both sequential and parallel convergent message-passing algorithms which have similar form to BP. The algorithms are based on a general message-passing architecture designed for a class of problems of the type $f(b) + \sum_i h_i(b)$ with $f(b)$ being strictly convex and $h_i$ being convex, continuous and proper. We have shown the basic steps of fitting the constrained convex free energy problem into this framework. We limited the discussions to factor graphs where the neighborhoods of every pair of factor nodes have at most a single intersection. This limitation can easily be removed by replacing the term $c_{i\alpha}(H(b_\alpha) - H(b_i))$ with the term $c_{\alpha,\beta}(H(b_\alpha) - H(b_\beta))$ for every pair of factor nodes. This replacement propagates mechanically into subsequent steps of the derivation — as would be found in a more detailed follow-up of this paper.

As for the second issue, we have proposed a heuristic principle where among all admissible parameters we choose the one most closest to the Bethe free energy (using Laplace principle of insufficient reasoning). Empirical results show that for certain graphs, like a grid, we obtain very close marginal results to those obtained by the TRW free energy. For random graphs we obtain a very different free energy from TRW and superior accuracy of marginal estimation. The results suggest that our heuristic for setting up the convex free energy satisfies what we were after, i.e., to get approximations similar to BP but in guaranteed (globally) convergent framework. Future work is required for obtaining a firmer theoretical understanding about the applicability of our heuristic and relation to TRW free energy in particular.

# References

Bertsekas, D., Nedić, A., & Ozdaglar, A. (2003). *Convex analysis and optimization*. Athena Scientific Belmont, Mass.

Bregman, L., Censor, Y., & Reich, S. (1999). Dykstras algorithm as the nonlinear extension of Bregmans optimization method. *Journal of Convex Analysis*, *6*, 319–333.

Dykstra, R. (1983). An Algorithm for Restricted Least Squares Regression. *Journal of the American Statistical Association*, *78*, 837–842.

Globerson, A., & Jaakkola, T. (2007a). Approximate inference using conditional entropy decompositions.

Globerson, A., & Jaakkola, T. (2007b). Convergent Propagation Algorithms via Oriented Trees. *Uncertainty in Artificial Intelligence (UAI 2007)*.

Heskes, T. (2004). On the Uniqueness of Loopy Belief Propagation Fixed Points. *Neural Computation*, *16*, 2379–2413.

Heskes, T. (2006). Convexity Arguments for Efficient Minimization of the Bethe and Kikuchi Free Energies. *Journal of Artificial Intelligence Research*, *26*, 153–190.

Kschischang, F., Frey, B., & Loeliger, H. (2001). Factor graphs and the sum-product algorithm. *IEEE Transactions on Information Theory*, *47*, 498–519.

Mooij, J.M., & Kappen, H.J. (2005). Sufficient conditions for convergence of loopy belief propagation. *Uncertainty in Artificial Intelligence (UAI 2005)*.

Murphy, K., Weiss, Y., & Jordan, M. (1999). Loopy belief propagation for approximate inference: An empirical study. *Proceedings of Uncertainty in AI*, 467–475.

Rockafellar, R. (1970). *Convex Analysis*. Princeton University Press.

Tseng, P. (1993). Dual coordinate ascent methods for non-strictly convex minimization. *Mathematical Programming*, *59*, 231–247.

von Neumann, J. (1950). Functional Operators, Vol. II: The Geometry of Orthogonal Spaces. *Annals of Math. Studies*, *22*.

Wainwright, M., Jaakkola, T., & Willsky, A. (2005). A new class of upper bounds on the log partition function. *Information Theory, IEEE Transactions on*, *51*, 2313–2335.

Weiss, Y., Yanover, C., & Meltzer, T. (2007). MAP Estimation, Linear Programming and Belief Propagation with Convex Free Energies. *Uncertainty in Artificial Intelligence (UAI 2007)*.

Yedidia, J., Freeman, W., & Weiss, Y. (2005). Constructing free-energy approximations and generalized belief propagation algorithms. *Information Theory, IEEE Transactions on*, *51*, 2282–2312.

Yuille, A. (2002). CCCP Algorithms to Minimize the Bethe and Kikuchi Free Energies: Convergent Alternatives to Belief Propagation. *Neural Computation*, *14*, 1691–1722.

## A  Sequential and Parallel Block Updates for $\min_b f(b) + \sum_i h_i(b)$

Recall the $f(b)$ is a strictly convex real-valued function and the functions $h_i$ are convex, proper and continuous. We quote below two basic theorems from convex duality (cf. (Bertsekas et al., 2003)) which we will use as building blocks for our algorithms.

**Theorem 1 Basic Fenchel Duality I**
*Let $g(\mathbf{b})$ be a convex and differentiable function and let $h(\mathbf{b})$ be a proper convex and continuous function, and let $h^*(\boldsymbol{\lambda}) = \max_{\mathbf{b}} \left\{ \mathbf{b}^\top \boldsymbol{\lambda} - h(\mathbf{b}) \right\}$ be its conjugate dual function. Consider the primal and dual programs:*

$$\text{Primal:} \quad \min_{\mathbf{b}} g(\mathbf{b}) + h(\mathbf{b})$$

$$\text{Dual:} \quad \max_{\boldsymbol{\lambda}} \left\{ \min_{\mathbf{b}} \left( g(\mathbf{b}) + \mathbf{b}^\top \boldsymbol{\lambda} \right) - h^*(\boldsymbol{\lambda}) \right\}$$

*then there is no duality gap and the optimal primal-dual pair $\mathbf{b}^*, \boldsymbol{\lambda}^*$ satisfies $\nabla g(\mathbf{b}^*) = -\boldsymbol{\lambda}^*$.*

The next theorem is a generalized version of the one above:

**Theorem 2 Basic Fenchel Duality II**
*Let $f(\mathbf{b})$ be a strictly convex and differentiable function and let $h_i(\mathbf{b})$ be proper convex and continuous functions, and let $h_i^*(\boldsymbol{\lambda}) = \max_{\mathbf{b}} \left\{ \mathbf{b}^\top \boldsymbol{\lambda} - h_i(\mathbf{b}) \right\}$ be their conjugate dual functions. Consider the primal and dual programs:*

$$\text{Primal:} \quad \min_{\mathbf{b}} f(\mathbf{b}) + \sum_{i=1}^{n} h_i(\mathbf{b})$$

$$\text{Dual:} \quad \max_{\boldsymbol{\lambda}_1,\ldots,\boldsymbol{\lambda}_n} \left\{ \min_{\mathbf{b}} \left( f(\mathbf{b}) + \mathbf{b}^\top \sum_{i=1}^{n} \boldsymbol{\lambda}_i \right) - \sum_{i=1}^{n} h_i^*(\boldsymbol{\lambda}_i) \right\}$$

*then there is no duality gap, and the optimal primal-dual pair $\mathbf{b}^*, \boldsymbol{\lambda}_i^*$ satisfies $\nabla f(\mathbf{b}^*) = -\sum_{i=1}^{n} \boldsymbol{\lambda}_i^*$.*

### A.1  The Sequential Block Update Algorithm

Since $f(\mathbf{b})$ is strictly convex, then its conjugate dual $\min_{\mathbf{b}} \left( f(\mathbf{b}) + \mathbf{b}^\top \sum_{i=1}^{n} \boldsymbol{\lambda}_i \right)$ is differentiable (see (Rockafellar, 1970)). In this case a block dual ascent optimization scheme converges to the global maxima (Tseng, 1993). Our algorithm alternates over $\boldsymbol{\lambda}_1, ..., \boldsymbol{\lambda}_n$ by optimizing $\boldsymbol{\lambda}_i$ while fixing $\boldsymbol{\lambda}_j$ for $j \neq i$.

Let $\boldsymbol{\mu}_i = \sum_{j \neq i} \boldsymbol{\lambda}_i$ and define the following dual algorithmic building block:

$$\max_{\boldsymbol{\lambda}_i} \left\{ \min_{\mathbf{b}} \left( f(\mathbf{b}) + \mathbf{b}^\top \boldsymbol{\mu}_i + \mathbf{b}^\top \boldsymbol{\lambda}_i \right) - h_i^*(\boldsymbol{\lambda}_i) \right\} \quad (8)$$

To recover $\boldsymbol{\lambda}_i$ one can use Theorem 1: Set $g(\mathbf{b}) \leftarrow f(\mathbf{b}) + \mathbf{b}^\top \boldsymbol{\mu}_i$ and $h(\mathbf{b}) \leftarrow h_i(\mathbf{b})$, and solve the primal program:

$$\mathbf{b}^* = \operatorname*{argmin}_{\mathbf{b} \in domain(h_i)} \left\{ f(\mathbf{b}) + \mathbf{b}^\top \boldsymbol{\mu}_i + h_i(\mathbf{b}) \right\}$$

From the Lagrange optimality condition of Theorem 1 we recover $\boldsymbol{\lambda}_i^*$

$$\boldsymbol{\lambda}_i^* = -\boldsymbol{\mu}_i - \nabla f(\mathbf{b}^*)$$

Taken together, one obtains Algorithm 1 described in Section 3.

### A.2 The Parallel Block Update Algorithm

We begin by stating and proving the following theorem:

**Theorem 3** *Let $f(\mathbf{b})$ be a strictly convex and differentiable function and let $h_i(\mathbf{b})$ be proper convex and continuous functions, and let $h_i^*(\boldsymbol{\lambda}) = \max_{\mathbf{b}} \left\{ \mathbf{b}^\top \boldsymbol{\lambda} - h_i(\mathbf{b}) \right\}$ be their conjugate dual functions. The following is a primal/dual pair with no duality-gap:*

$$(P) \min_{\mathbf{b}} f(\mathbf{b}) + \sum_{i=1}^n h_i(\mathbf{b})$$

$$(D) \max_{\substack{\boldsymbol{\lambda}_1, \ldots, \boldsymbol{\lambda}_n \\ \sum_{i=1}^n \boldsymbol{\mu}_i = 0}} \left\{ \sum_{i=1}^n \left( \min_{\mathbf{b}} \left( \frac{1}{n} f(\mathbf{b}) + \mathbf{b}^\top (\boldsymbol{\lambda}_i + \boldsymbol{\mu}_i) \right) - h_i^*(\boldsymbol{\lambda}_i) \right) \right\}$$

*Furthermore, the optimal primal-dual pair $\mathbf{b}^*, \boldsymbol{\lambda}_i^*$ satisfies $\nabla f(\mathbf{b}^*) = -\sum_{i=1}^n \boldsymbol{\lambda}_i^*$.*

**Proof:** we introduce an equivalent primal function $\sum_{i=1}^n \left( \frac{1}{n} f(\mathbf{b}_i) + h_i(\mathbf{y}_i) \right)$ subject to the constraints $\mathbf{b} = \mathbf{b}_i$ and $\mathbf{b}_i = \mathbf{y}_i$ for every $i$. The Lagrangian $L(\mathbf{b}, \mathbf{y}_i, \boldsymbol{\lambda}_i, \boldsymbol{\mu}_i)$ and the Lagrange dual function $q(\boldsymbol{\lambda}_i, \boldsymbol{\mu}_i) = \min_{\mathbf{b}, \mathbf{b}_i, \mathbf{y}_i} L()$ take the form:

$$L(\cdot) = \sum_{i=1}^n \left( \frac{1}{n} f(\mathbf{b}_i) + h_i(\mathbf{y}_i) + \boldsymbol{\mu}_i^\top (\mathbf{b}_i - \mathbf{b}) + \boldsymbol{\lambda}_i^\top (\mathbf{b}_i - \mathbf{y}_i) \right)$$

$$q() = \sum_{i=1}^n \left( \min_{\mathbf{b}} \left( \frac{1}{n} f(\mathbf{b}) + \mathbf{b}^\top (\boldsymbol{\lambda}_i + \boldsymbol{\mu}_i) \right) - h_i^*(\boldsymbol{\lambda}_i) \right)$$

Note that whenever $\sum_{i=1}^n \boldsymbol{\mu}_i \neq 0$ the dual function attains the value $q() = -\infty$. Since we seek to maximize the dual function we need to optimize $\boldsymbol{\mu}_i$ in its domain, i.e. $\sum_{i=1}^n \boldsymbol{\mu}_i = 0$. □

The function $\sum_{i=1}^n \frac{1}{n} f(\mathbf{b}_i)$ is strictly convex therefore its conjugate dual is differentiable (see (Rockafellar, 1970)). In this case a block dual ascent optimization scheme converges to the global maxima (Tseng, 1993). Our algorithm alternates through optimizing $\boldsymbol{\lambda}_i$ (in parallel) while fixing $\boldsymbol{\mu}_i$ followed by optimizing $\boldsymbol{\mu}_i$ (by a closed form solution) while fixing $\boldsymbol{\lambda}_i$. We formulate our dual algorithmic building block with respect to $\boldsymbol{\lambda}_i$ using Theorem 1: Set $g(\mathbf{b}) \leftarrow \frac{1}{n} f(\mathbf{b}) + \mathbf{b}^\top \boldsymbol{\mu}_i$ and $h(\mathbf{b}) \leftarrow h_i(\mathbf{b})$, and solve the primal program:

$$\mathbf{b}^* = \operatorname*{argmin}_{\mathbf{b} \in domain(h_i)} \left\{ \frac{1}{n} f(\mathbf{b}) + \mathbf{b}^\top \boldsymbol{\mu}_i + h_i(\mathbf{b}) \right\}$$

From Lagrange optimality condition in Theorem 1 we recover $\boldsymbol{\lambda}_i^*$

$$\boldsymbol{\lambda}_i^* = -\boldsymbol{\mu}_i - \frac{1}{n} \nabla f(\mathbf{b}^*) \quad (9)$$

We turn to find the closed-form solution for optimizing $\boldsymbol{\mu}_1, \ldots, \boldsymbol{\mu}_n$ while fixing $\boldsymbol{\lambda}_1, \ldots, \boldsymbol{\lambda}_n$ using Theorem 1: Set $g(\mathbf{b}_1, \ldots, \mathbf{b}_n) \leftarrow \frac{1}{n} \sum_{i=1}^n (f(\mathbf{b}_i) + \mathbf{b}_i^\top \boldsymbol{\lambda}_i)$ and set $h(\mathbf{b}_1, \ldots, \mathbf{b}_n)$ to be the indicator function that attains the value zero if $\mathbf{b}_1 = \cdots = \mathbf{b}_n$ and infinity otherwise. The conjugate function of $h(\mathbf{b}_1, \ldots, \mathbf{b}_n)$ is the indicator function $h^*(\boldsymbol{\mu}_1, \ldots, \boldsymbol{\mu}_n)$ whose value is zero if $\sum_{i=1}^n \boldsymbol{\mu}_i = 0$ and $\infty$ otherwise. The primal program:

$$\operatorname*{argmin}_{\mathbf{b}_1, \ldots, \mathbf{b}_n \in domain(h)} \left\{ \sum_{i=1}^n \left( \frac{1}{n} f(\mathbf{b}_i) + \mathbf{b}_i^\top \boldsymbol{\lambda}_i \right) \right\}$$

can be further simplified by taking into account the domain of $h(\mathbf{b}_1, \ldots, \mathbf{b}_n)$, i.e. restricting all the $\mathbf{b}_i$ to equal some vector $\mathbf{b} \in \mathbb{R}^n$:

$$\operatorname*{argmin}_{\mathbf{b} \in \mathbb{R}^n} \left\{ f(\mathbf{b}) + \mathbf{b}^\top \sum_{i=1}^n \boldsymbol{\lambda}_i \right\}$$

Since $f(\mathbf{b})$ is real-valued function and the optimization is unconstrained the optimal vector $\mathbf{b}^*$ satisfies $\nabla f(\mathbf{b}^*) = -\sum_{i=1}^n \boldsymbol{\lambda}_i$.

Theorem 1 asserts the Lagrange multipliers $\boldsymbol{\nu}^* = \boldsymbol{\mu}_1^*, \ldots, \boldsymbol{\mu}_n^*$ equals the gradient of $g(\mathbf{b}_1^*, \ldots, \mathbf{b}_n^*)$, or equivalently $\boldsymbol{\mu}_i^* = -\frac{1}{n} \nabla f(\mathbf{b}_i^*) - \boldsymbol{\lambda}_i$. In the preceding paragraph we argued that $\nabla f(\mathbf{b}^*) = -\sum_{i=1}^n \boldsymbol{\lambda}_i$ so we derive the update rule for $\boldsymbol{\mu}_i^*$

$$\boldsymbol{\mu}_i^* = -\boldsymbol{\lambda}_i + \frac{1}{n} \sum_{i=1}^n \boldsymbol{\lambda}_i \quad (10)$$

Taken together, one obtains Algorithm 2 described in Section 3.